\documentclass[conference]{IEEEtran}
\IEEEoverridecommandlockouts
\usepackage{cite}
\usepackage{amsmath,amssymb,amsfonts}
\usepackage{graphicx}
\usepackage{textcomp}
\usepackage{xcolor}
\usepackage{algorithm}
\usepackage{tabularx} 
\usepackage[noend]{algpseudocode}

\def\BibTeX{{\rm B\kern-.05em{\sc i\kern-.025em b}\kern-.08em
    T\kern-.1667em\lower.7ex\hbox{E}\kern-.125emX}}
\begin{document}
\makeatletter
\newcommand{\newlineauthors}{%
  \end{@IEEEauthorhalign}\hfill\mbox{}\par
  \mbox{}\hfill\begin{@IEEEauthorhalign}
}
\makeatother
\title{Efficient image retrieval using multi neural hash codes and bloom filters}

\author{
\IEEEauthorblockN{Sourin Chakrabarti}
\IEEEauthorblockA{\textit{iit2016513@iiita.ac.in} \\
\textit{IIIT Allahabad, India}}
}
\IEEEoverridecommandlockouts
\IEEEpubid{\makebox[\columnwidth]{978-1-5386-5541-2/18/\$31.00~\copyright2018 IEEE \hfill} \hspace{\columnsep}\makebox[\columnwidth]{ }}
\maketitle

\begin{abstract}
This paper aims to deliver an efficient and modified approach for image retrieval using multiple neural hash codes and limiting the number of queries using bloom filters by identifying false positives beforehand. Traditional approaches involving neural networks for image retrieval tasks tend to use higher layers for feature extraction. But it has been seen that the activations of lower layers have proven to be more effective in a number of scenarios. In our approach, we have leveraged the use of local deep convolutional neural networks which combines the powers of both the features of lower and higher layers for creating feature maps which are then compressed using PCA and fed to a bloom filter after binary sequencing using a modified multi k-means approach. The feature maps obtained are further used in the image retrieval process in a hierarchical coarse-to-fine manner by first comparing the images in the higher layers for semantically similar images and then gradually moving towards the lower layers searching for structural similarities. While searching, the neural hashes for the query image are again calculated and queried in the bloom filter which tells us whether the query image is absent in the set or maybe present. If the bloom filter doesn't necessarily rule out the query, then it goes into the image retrieval process. This approach can be particularly helpful in cases where the image store is distributed since the approach supports parallel querying.
\end{abstract}


\section{Introduction}
Convolutional neural networks(CNN) \cite{b30} have proven to be very useful in image classification and identification tasks. Apart from these, an active research topic in recent times has been image retrieval. Often it is seen that the search space is quite big and it takes lots of time for an exhaustive search through the complete database. Hence, lots of recent studies have also been focused on reducing the number of queries to the database using various data structures that have minimal knowledge of the content of the images but enough for them to be useful for a pre-query optimization along with the image retrieval task.

 The approach followed in this paper mainly revolves around the success of CNNs in extracting features from images \cite{b31}. The final output of the CNN just before the layer for classification contains the final feature map required for classification. This property has been often leveraged by various image descriptors to produce a compressed representation of the images. But interestingly enough, on comparing the performance of the values of the intermediate layers with the final layer for retrieval, it was seen that often the best results were obtained out of the intermediate layers \cite{b2}. Our approach makes use of all the information available from the image by using neural hash codes from both the lower and higher layers. 

It is seen that neural codes have performed particularly well when it comes to image retrieval tasks \cite{b2,b17}. Using CNN image descriptors along with bloom filters have been tried before by \cite{b1} to increase search efficiency. We propose a unique approach of using neural codes from multiple layers simultaneously for image retrieval tasks and pre-query filtering tasks. For the generation of the neural codes, dimensionality reduction was another area that needed consideration. Since PCA compression provides a good degradation to performance trade-off \cite{b2}, We use it to compress the layer outputs. The final number of dimensions was fixed to 128 since it provided a negligible loss of accuracy. The compressed feature vectors were converted to binary sequences using a modification of the multi k-means approach to accommodate multiple feature vectors and then fed to the bloom filter to limit queries. For image retrieval, the feature vectors of the higher layers helped in similarity matching in the coarse stage whereas lower layers gave more insights about the fine structural details of the image and hence were used for further retrieval.

Finally, experiments on the Oxford 5k \cite{b27}, INRIA holidays \cite{b29} and Paris 6k \cite{b28} data sets were carried out to report the precision of our proposed method and the time consumed for the queries. We also show the effect on precision caused by varying the number of layers used for generating neural codes for various data sets along with varying the size of the bloom filter.
\IEEEpubidadjcol

\section{Related Work}
Deep learning-based image descriptors have been researched for long. SIFT-based descriptors \cite{b26} were widely used for a long for image description tasks before the inception of other more robust and efficient descriptors such as VLAD \cite{b24} and Fischer descriptors \cite{b33}. Moreover, recently triangulation embedding along with democratic aggregation has been shown to outperform Fischer vectors \cite{b16}. CNN based image descriptors have been known to outperform the above descriptors on numerous occasions and have proven to act as a baseline in image recognition and retrieval tasks \cite{b17}. Often CNNs are fine-tuned to work for image retrieval tasks. \cite{b7} proposed a fully automated procedure for fine-tuning a CNN for image retrieval. \cite{b2} uses lower layers of the CNN for image description tasks and compares it with higher layers showing that often the intermediate layers contain more valuable information required for image description. Capturing local CNN features and compression using VLAD has been experimented by \cite{b4}. Accumulation of these deep features has been discussed in \cite{b6}. On the other hand, \cite{b15} proposed the extraction of image features through capturing them from various parts of the image through the construction of windows(more like an R-CNN) and finally compressing using Fischer vectors or PCA compression. These techniques also find heavy usage in frame retrieval from large videos as depicted by \cite{b12}.

Output produced by CNNs or any image descriptor is generally of larger dimensions than required. These larger dimensions need to be reduced in order for the image retrieval systems to be efficient. This reduction can be achieved by adding hidden layers before the final classification layers in the case of CNNs. Some of the most prominent techniques employed in this field are shown in \cite{b34}. \cite{b35} suggested using the image features and the ground truth texts to devise a similarity graph and hence learn relations among the elements of the data set. In \cite{b36}, neural codes from a CNN were generated by first pre-training on an Image-net following which another layer was added to the network which generated the required hashes and finally retrieval was carried out using hierarchical search using both the hash codes and the features of certain layers. \cite{b14} used multi-label images to devise a new hashing approach. They used CNNs for finding features and then used a fully connected layer along with a loss function based on the multi-label supervision to generate the hashes and learn the features at the same time. \cite{b1} also used CNN descriptors with hashing and indexing along with bloom filters on sharded databases to reduce the number of queries for image retrieval.

Image matching algorithms are quite important in an image retrieval process. Traditionally, image matching problems were re-framed as graph optimization problems. Several graph-based matching algorithms based on Markov Random fields were suggested by \cite{b37,b38}. Using a coarse to fine approach to establish dense pixel-level correspondences using a randomized search was suggested by \cite{b40}. WarpNet \cite{b39} proposed an architecture that could match objects in various images. \cite{b3} suggested using a deep learning based hierarchical image matching architecture which proved to be quite suitable for our cause. 

The features obtained were typically hashed and stored in inverted file systems. \cite{b44} introduced the inverted multi-index which replaces standard quantization with product quantization. \cite{b5} further improved upon this approach by random initialization of a k-means dictionary and storing the results in tries.

Bloom filters since their inception \cite{b19} have proved to be quite effective in identifying true negatives in searching operations. But their application in fields of image processing has been limited. \cite{b18} proposed Bloomier filters for storing features of an image data set which made them more memory efficient than storing them in hashes. Image descriptors have found themselves together with bloom filters on numerous occasions such as in \cite{b1} where bloom filters were used to construct a feature descriptor that used hash functions that differentiate between inputs of different categories. Applications such as image retrieval from videos which include memory-intensive tasks have employed the use of bloom filters along with novel image descriptors to improve efficiency \cite{b12}. Bloom filters have also performed well as image descriptors in the past \cite{b41}. Recent advances in fields of bloom filters have suggested that bloom filters can be modified to work with heavy throughput of data with acceptable accuracy like the case with neural bloom filters \cite{b8}.

\section{Proposed methodology}
Our proposed methodology for efficient image retrieval primarily revolves around training the model used, identifying the lower layers to be used for feature extraction, feature compression, binary sequencing, insertion in the bloom filter and database and then finally image retrieval.
\subsection{CNN architecture}

In our setup, we leverage the model used by \cite{b2} to successfully extract neural codes from lower layers. We use our own modification of the system to include neural codes from multiple layers in one feed forward cycle. The model consists of 5 primary units. A unit consists of a convolutional layer followed by a pooling layer that employs max pooling and then finally an activation layer using the non-linear ReLu transform. These 5 units are followed by 3 fully connected layers at the end of which the output is received in one hot encoding. The architecture is shown in Figure \ref{fig0}. While training the model softmax loss function was used. The model was pretrained on the Image-Net \cite{b43} classes and fine-tuned by training on the relevant datasets for our use case.
\begin{figure*}[!htb]
\includegraphics[width=\textwidth]{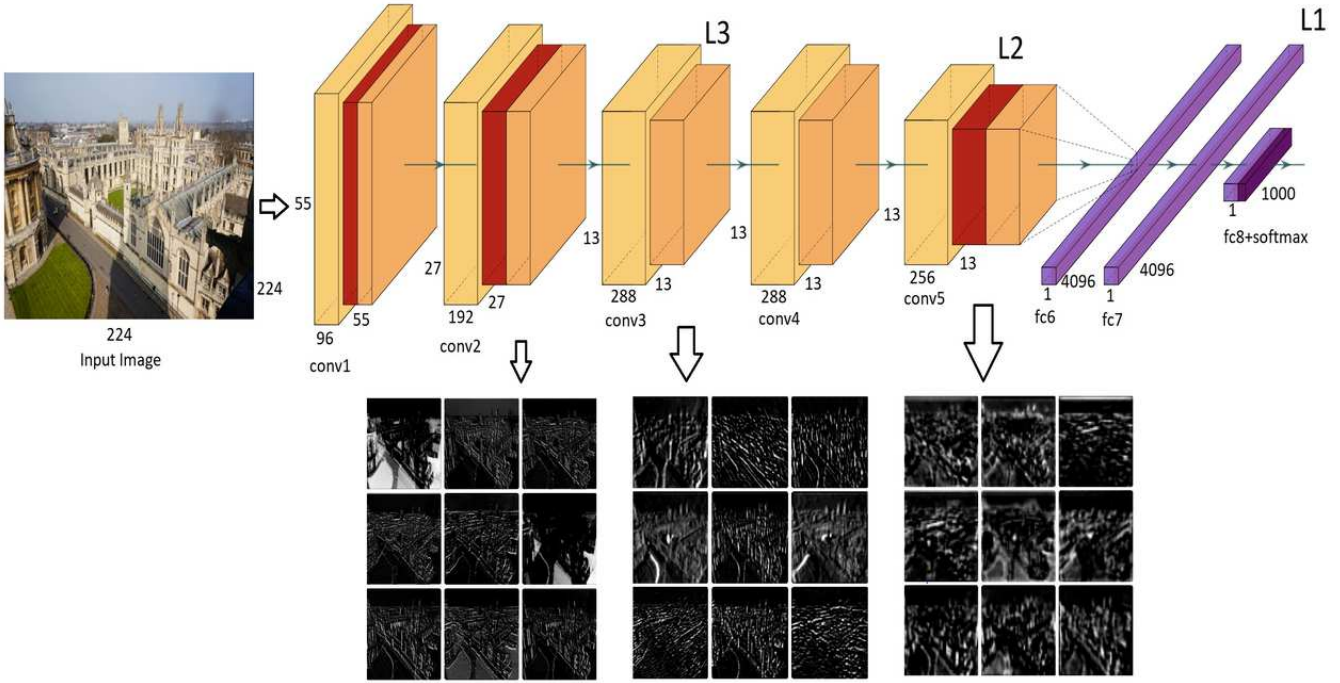}
\caption{The CNN architecture used for training. Yellow slabs
correspond to convolution layers, red slabs correspond to the
max pooling, and the orange slabs correspond to the ReLu layers. Layers fc6, fc7, and fc8 are fully connected to the preceding
layers. The layers used for neural codes are L1, L2 and L3. The activations of intermediate layers shown are randomly chosen from all the available filters for the layer.} \label{fig0}
\end{figure*}

For benchmarking, images from the following datasets were used for training and testing purposes:
\paragraph{Oxford 5k}
The Oxford buildings dataset consists of a collection of 5062 images spanning over the colleges of Oxford. The ground truth text for each one refers to one of the 11 landmark buildings. For each image, labels were associated with them which gave information about the quality of the image. About 55 query images are present with each one belonging to either of the 11 landmarks.
\paragraph{Paris 6k}
The Paris dataset consists of a collection of 6412 images spanning over the landmarks of Paris. The ground truth text for each one refers to one of the 12 landmark buildings.
\paragraph{INRIA holidays}
This dataset include images from holiday destination containing a wide array of images and corresponding classes. It consists of 991 training and 500 query images spanning across 500 classes. Certain images were rotated but our model being rotation invariant, it didn't make a difference.

The images were pre processed before feeding them to the neural network. All the images were resized to 224x224 to be provided as input. For training, the strides used were 4 for all the layers except the first and 1 for the first layer. The model was pretrained on Image-Net classes but was again trained using the datasets mentioned above to suit our use-case. Once the model was trained, randomly selected training images were passed through the neural network as features from layers L1, L2 and L3 were collected to be introduced into the bloom filter. Once the feature maps were collected, PCA compression was applied on the maps to compress the feature vector to a size of 128. This length was decided upon due to the size of our test dataset.
\subsection{Binary sequencing}
Our approach uses a modified version of the proposed by \cite{b5}. They randomly initialized a k-means dictionary with some feature vectors from the training dataset. Now, whenever an image is introduced, its similarity is compared with each of the current centroids. This computation is done by considering the L2 distances of all the pairs of feature vectors in our case. Once the L2 distances are computed, they are sorted and distances with centroids below a certain threshold are considered. Once an image is included in a cluster, the corresponding bit is marked as 1 in the generated binary sequence for that image and the rest are marked 0. A feature can be assigned to more than one centroids.
\subsection{Hashing and Bloom filters}
The binary sequenced codes obtained from the three layers were hashed via a hash function. The choice of hash function for a bloom filter should conform to uniformity and should be logically independent. Over the years various hash functions have found applications to bloom filters, but the primary ones used along with binary signatures include the Murmur3 \cite{b42}, cryptographic hashes like SHA-256, LCGs etc. 

A bloom filter essentially is a data structure that probabilistically determines the presence of an element in a set. False positive outputs are deterministically given by a bloom filter. In traditional bloom filters, k hash functions are used to compute k hashes of an item to be inserted into the structure. Each of the k hashes return a position to be marked in the filter. During retrieval, the query element is again hashed against all the hash functions and the returned positions are matched from the filter. If any one of the positions is unset, it is inferred that the element is definitely not present in the original data set. Whereas if all the positions are marked, then we can't conclude anything deterministically.

The probability of presence of a false positive element in a bloom filter is given by the equation :
\begin{equation}
P = (1 - (1-\frac{1}{m})^{nk})^k
\end{equation}
where n is the size of the data set, m is the size of the bloom filter and k is the number of hash functions used \cite{b19}. We can obtain optimal values of m for a fixed k and n by the equation:
\begin{equation}
m = kn\ln{2} 
\end{equation}
In our case, we have k ranging from 1 to 3. Hence, the corresponding bloom filter sizes would be ideally from 1.5n to 4.5n approximately.
To facilitate our bloom filter, we have used Murmur3 hash function to compute hashes from the binary sequences from the layers L1, L2 and L3. Once the hashes are computed, they are inserted into the bloom filters of the fixed size. Once this process is completed, the query images are used for testing.
\subsection{Storage}
The compressed feature vectors were stored in a hierarchical manner with the L3 layer at the top followed by the L2 and the L1 layer. This model allowed more layers to be used in the matching process in later developments. The average cosine distance between images of the same class were computed during training to be used as a threshold during retrieval process. 
\subsection{Retrieval process}
For the limiting process, the compressed feature vectors are again extracted from the CNN layers by passing the query image through the neural network. The compressed feature vectors are quantized and passed to the bloom filter post hashing. If atleast one bit is unset, then the image retrieval process must stop. In case the filter couldn't give certainty, then the L3 layer features are compared initially. The comparison is carried out by finding the nearest neighbours based on the cosine distances of two feature vectors and a threshold as calculated previously. Once, we get a refined search space, the process is repeated with finer thresholds for the L2 layer search vectors and finally the best structural matches are found by comparing the L3 layer feature vectors.

The complete algorithm is briefly represented by Figure \ref{fig1}.
\begin{figure}[!htb]
\includegraphics[width=0.5\textwidth]{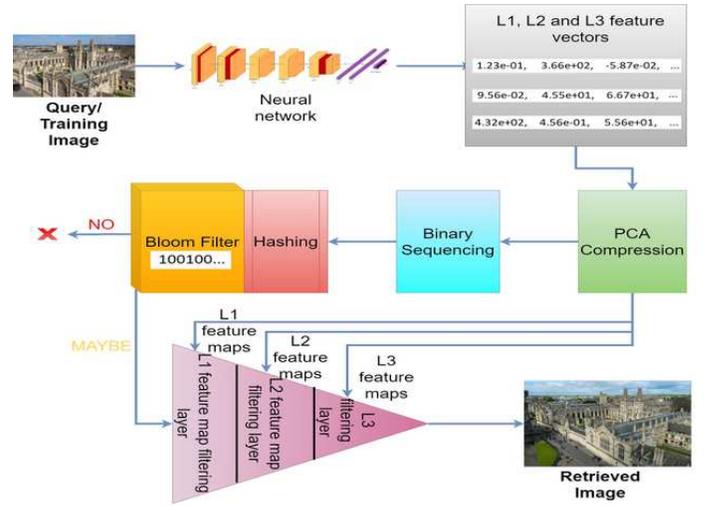}
\caption{Complete architecture: The query/training image is fed to the neural network for neural code generation which are PCA compressed. These compressed vectors are sent to the respective storage layers for retrieval. The compressed vectors are then binary sequenced and hashed before feeding them to the bloom filter while training or searching them in the bloom filter while querying. Depending on the output of the bloom filter, the query proceeds. For retrieval, PCA compressed vector from the query is first compared with the L1 feature maps followed by L2 and L3 before providing a final output.} \label{fig1}
\end{figure}
\section {Experimentation and Results}
The proposed methodology was tested for mean average precision using the Oxford 5k, Paris 6k and the Holidays dataset. Also, distractor images from the Flickr 100 dataset were used to test the robustness of the system with garbage data. Final feature maps were of the size 128 since they proved to be a good agreement between efficiency and accuracy \cite{b2}. The final reported accuracy was based on the mean average precision(mAP) for the system.

The first setup consisted of training the network using only the dataset images. Tests on every dataset were carried out first using only layer L1 as the feature vector followed by L1 and L2 and finally all the three. The size of the bloom filter was set to be 2n and 5n, n being the number of data items in the bloom filter. The threshold for binary sequencing was fixed to be 10 and the number of centroids was fixed to be 64. It was seen that as the number of feature vectors used for input increased, the accuracy increased generally. Moreover, this increase in accuracy was also seen as the size of the filter was increased. We see that the performance was particularly good as compared to the current state-of-the-art with a very minimal compute time. The time required for computation generally decreased with the increase in the size of the filters and also with the increase in the number of hashes used.

\begin{table}
\caption{This table contains information about the mAP and the average time(in s) for the experiments carried out on the Oxford 5k, Paris 6k and the Holidays dataset without any distractor image. n is the number of images stored in the database. L1, L2 and L3 are the layers of feature vectors. The threshold for binary sequencing was fixed at 10 for a 64 bit sequence.}\label{tab1}
\begin{center}
\begin{tabular}{| m{6em} | m{0.7cm}| m{0.7cm} |  m{0.7cm} |  m{0.7cm} |  m{0.7cm} |  m{0.7cm} | }
\hline
\bfseries Dataset &  \multicolumn{2}{|c|}{\bfseries Oxford 5k} & \multicolumn{2}{|c|}{\bfseries Paris 6k} & \multicolumn{2}{|c|}{\bfseries Holidays}\\
\hline
\bfseries Filter size & \bfseries 2n & \bfseries 5n & \bfseries 2n & \bfseries 5n & \bfseries 2n & \bfseries 5n \\
\hline
L1 & 40.31 & 41.34 & 58.73 & 60.01 & 69.54 & 70.96\\
& 1.48 & 1.21 & 1.97 & 1.55 & 30.60 & 27.04\\
\hline
L1 \& L2 & 41.26 & 43.43 & 63.68 & 64.09 & 73.41 & 73.99\\
& 1.26 & 0.98 & 1.37 & 1.26 & 28.45 & 23.67\\
\hline
L1 \& L2 \& L3 & 46.68 & 49.88 & 65.33 & 67.79 & 76.02 & 77.34\\
& 1.34 & 1.15 & 1.19 & 1.13 & 23.26 & 20.89\\
\hline
\end{tabular}
\end{center}
\end{table}

\begin{figure}
\includegraphics[width=0.5\textwidth]{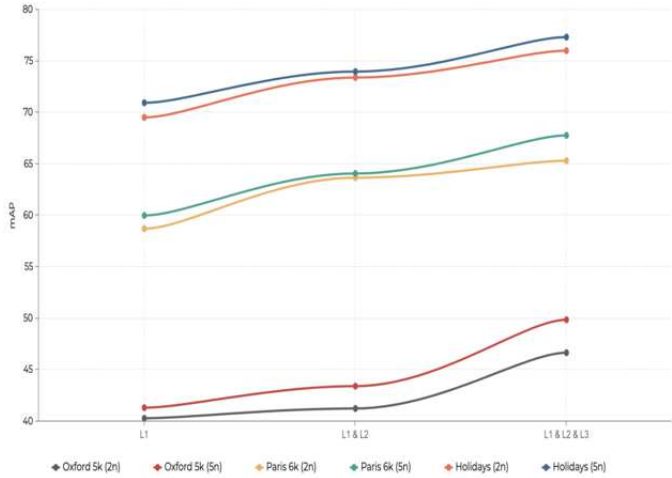}
\caption{This graph presents a visual representation of the data presented in Table \ref{tab1}. The horizontal axis represents the number of feature vectors whereas the vertical axis corresponds to the mAP.} \label{gr1}
\end{figure}

The second setup included mixing the datasets along with the distractor images. All three datasets were mixed and the model was trained on all the training classes. Tests were carried out simultaneously using bloom filter size twice and five times the sizes of the input set.

The test results were almost consistent with before. The accuracy was seen to dip significantly on using a single descriptor but it performed considerably well when all the descriptors were used. The overall speed of execution was reasonably faster due to a large number of false positives which got identified during early iterations.

\begin{table}
\caption{This table contains information about the mAP and the average time(in s) for the experiments carried out on the Oxford 5k, Paris 6k and the Holidays datasets mixed with the distractor images. n is the number of images stored in the database. L1, L2 and L3 are the layers of feature vectors. The threshold for binary sequencing was fixed at 10 for a 64 bit sequence.}\label{tab2}
\begin{center}
\begin{tabular}{| m{6em} | m{0.7cm}| m{0.7cm} |  m{0.7cm} |  m{0.7cm} |  m{0.7cm} |  m{0.7cm} | } 
\hline
\bfseries Dataset &  \multicolumn{2}{|c|}{\bfseries Oxford 5k} & \multicolumn{2}{|c|}{\bfseries Paris 6k} & \multicolumn{2}{|c|}{\bfseries Holidays}\\
\hline
\bfseries Filter size & \bfseries 2n & \bfseries 5n & \bfseries 2n & \bfseries 5n & \bfseries 2n & \bfseries 5n \\
\hline
L1 & 39.81 & 40.40 & 56.97 & 57.45 & 44.50 & 48.95\\
& 2.89 & 1.58 & 4.36 & 2.98 & 42.60 & 40.76\\
\hline
L1 \& L2 & 40.98 & 41.32 & 57.75 & 58.28 & 51.48 & 56.99\\
& 1.67 & 1.19 & 3.74 & 2.12 & 39.87 & 34.49\\
\hline
L1 \& L2 \& L3 & 41.67 & 42.89 & 58.13 & 60.98 & 58.86 & 60.07\\
& 1.42 & 1.27 & 2.96 & 1.92 & 35.21 & 33.47\\
\hline
\end{tabular}
\end{center}
\end{table}
\begin{figure}
\includegraphics[width=0.5\textwidth]{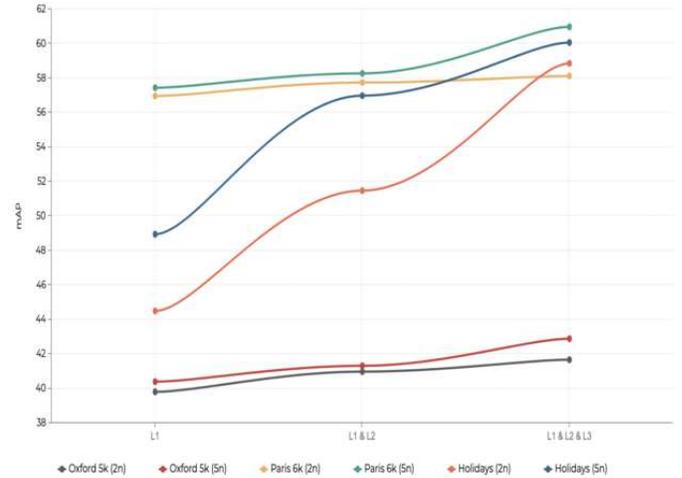}
\caption{This graph presents a visual representation of the data presented in Table \ref{tab2}. The horizontal axis represents the number of feature vectors whereas the vertical axis corresponds to the mAP.} \label{gr2}
\end{figure}

\section{Conclusion}
In this paper, we present a fresh approach for efficient image retrieval by using multiple neural hash codes which are PCA compressed and binary sequenced before feeding them to a bloom filter to reduce the average number of queries. The feature vectors are also used in the retrieval process which follows a hierarchical coarse-to-fine approach by using the higher layers for coarse search and the lower layers for a finer search. The followed approach shows precisions matching state-of-the-art while reducing the number of queries and hence mean query times greatly.

Further improvements made to algorithm can be towards trying out other state-of-the-art networks with a similar approach or experimenting with different feature compression processes. Modern image matching processes suited with multiple feature vectors can also be experimented with to increase accuracy.

\end{document}